\documentclass{article}

\usepackage{arxiv}

\usepackage[utf8]{inputenc} % allow utf-8 input
\usepackage[T1]{fontenc}    % use 8-bit T1 fonts
\usepackage{hyperref}       % hyperlinks
\usepackage{url}            % simple URL typesetting
\usepackage{booktabs}       % professional-quality tables
\usepackage{amsfonts}       % blackboard math symbols
\usepackage{nicefrac}       % compact symbols for 1/2, etc.
\usepackage{microtype}      % microtypography
\usepackage{graphicx}
\usepackage{natbib}
\usepackage{doi}
\usepackage{subcaption}
\usepackage{fontawesome5}
\usepackage{float}

\title{Modelling controllers for Cyber Physical Systems using Neural Networks}

\author{ \href{https://orcid.org/0000-0001-7993-7979}{\includegraphics[scale=0.06]{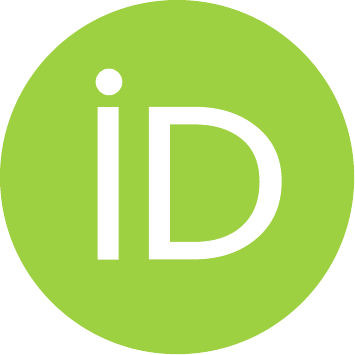}\hspace{1mm}Aravindakumar Vijayasri Mohan Kumar} \\
	Department of Computer Science\\
	University of Colorado Boulder\\
	Boulder, CO 80310 \\
	\texttt{arvi7401@colorado.edu} \\
}

\renewcommand{\shorttitle}{Independent Study Report - Fall 2022}

\hypersetup{
pdftitle={Modelling controllers for Cyber Physical Systems using Neural Networks},
pdfsubject={q-bio.NC, q-bio.QM},
pdfauthor={Aravindakumar Vijayasri Mohan Kumar},
pdfkeywords={Imitiation learning, Neural networks, Dagger},
}

\begin{document}
\maketitle

\begin{abstract}
Model Predictive Controllers (MPC) are widely used for controlling cyber-physical systems. It is an iterative process of optimizing the
prediction of the future states of a robot over a fixed time horizon. MPCs are effective in practice, but because they are computationally expensive and slow, they are not well suited for use in real-time applications. Overcoming the flaw can be accomplished by approximating an MPC's functionality. Neural networks are very good function approximators and are faster compared to an MPC.  It can be challenging to apply neural networks to control-based applications since the data does not match the i.i.d assumption. This study investigates various imitation learning methods for using a neural network in a control-based environment and evaluates their benefits and shortcomings.
\end{abstract}

% keywords can be removed

\section{Related Work}
Numerous studies concentrate on employing neural networks to approximate existing controller functions. \cite{phvalue} controls the pH levels in a laboratory-scale neutralization reactor using a neural network approximator instead of a PI controller. The authors of  \cite{hvac} control residential HVAC (Heating, Ventilation, and Air-Conditioning) systems using MPCs that are based on Artificial Neural Networks.  

In \cite{claviere2019trajectory}, the authors adopt a counter-example guided strategy for training neural networks. The initial data is derived from an MPC simulation. A network is trained using the initial data and then analyzed using a falsifier which generates counter-examples that form the additional training data for consequent cycles. This work adopts the notion of using MPC as a teacher for a neural network training and examines two imitation learning (\cite{imitation-learning}) approaches namely Behavior Cloning and Dataset Aggregation (Dagger) (\cite{dagger-paper}).

\section{Behavioral Cloning}
\begin{figure}[ht]
    \includegraphics[width=7cm]
    {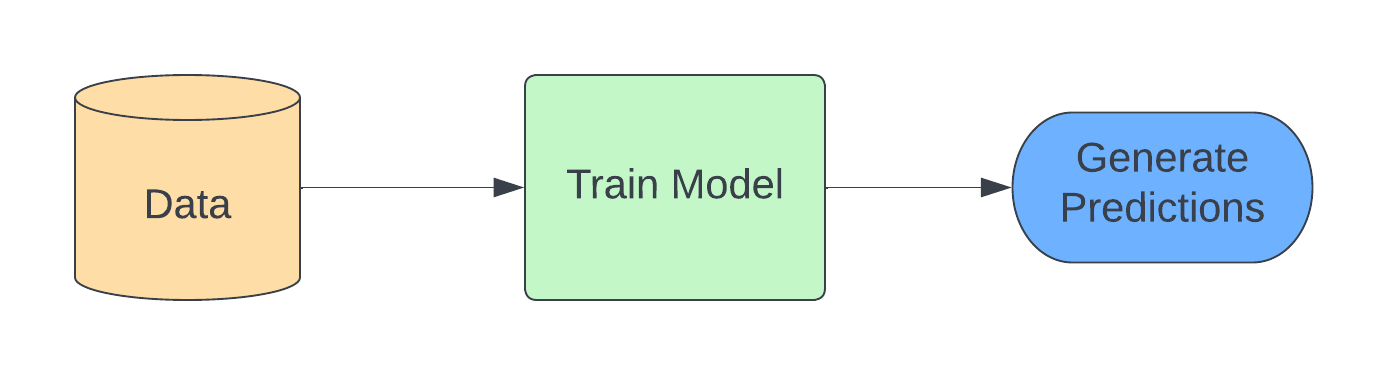}
    \centering
    \caption{Behavioral Cloning}
    \label{fig:BehavioralCloning}
\end{figure}

Behavioral cloning is a supervised learning approach where we train a neural network model with the data obtained by generating sample trajectories using an MPC controller. Fig \ref{fig:BehavioralCloning} depicts the process of the behavioral cloning approach. 

The main drawback of the behavioral cloning approach is the distributional shift that occurs between the learned policy distribution \(P_{\pi_{\theta}}\) and the original data distribution \(P_{data}\) which causes the error to compound after a single incorrect prediction.

\section{Dagger Approach}
Dagger is an imitation learning approach that tries to minimize the distributional shift between 
\(P_{\pi_{\theta}}\) and \(P_{data}\). Dagger approach iteratively collects data from \(P_{\pi_{\theta}}\) rather than from \(P_{data}\). 

\begin{enumerate}
    \item Train policy \(\pi_{\theta}\) from initial data \(D_{0}\) obtained from MPC controller.
    \item Run policy \(\pi_{\theta}\) to get new states \(S_{\pi}\).
    \item Input the states \(S_{\pi}\) to an MPC controller to get corresponding controls \(C_{\pi}\).
    \item These states and controls form the new data \(D_{\pi} = (S_{\pi}, C_{\pi})\).
    \item Aggregate \(D_{0}\) and \(D_{\pi}\) to get data for next iteration ie.,( \(D_{1} = D_{0} \cup D_{\pi}\))
\end{enumerate}

The intuition is that when we repeat the above procedure for several iterations the samples collected from \(P_{\pi_{\theta}}\) will dominate the original samples collected from \(P_{data}\) which in turn will minimize make 
\(P_{\pi_{\theta}} \approx P_{data}\). Here, MPC is used during every iteration of the training phase and acts as a teacher.

\section{Experiments}
\subsection{Plant model}
A bicycle robot model with 4 states and 2 controls is used as the plant model for this study. The description of the states and controls are as follows: 
\begin{enumerate}
    \item Four states: 
        \begin{itemize}
            \item y - position of the robot 
            \item v - velocity of the robot
            \item \(\theta\) - rotational position
            \item \(\gamma\) - tangent to the angle between the front and rear axels
        \end{itemize}
    \item Two controls:
        \begin{itemize}
            \item \(u_1\) - acceleration of the robot
            \item \(u_2\) - angular velocity of the robot
        \end{itemize}
\end{enumerate}
Here, \(y, v \in (-2, 2)\) and \(\theta, \gamma \in (-1, 1)\) are the bounds for the states and \(u_1, u_2 \in (-10, 10)\) are the bounds on the controls.

\subsection{Neural architecture}
\begin{figure}[ht]
    \includegraphics[width=7cm]{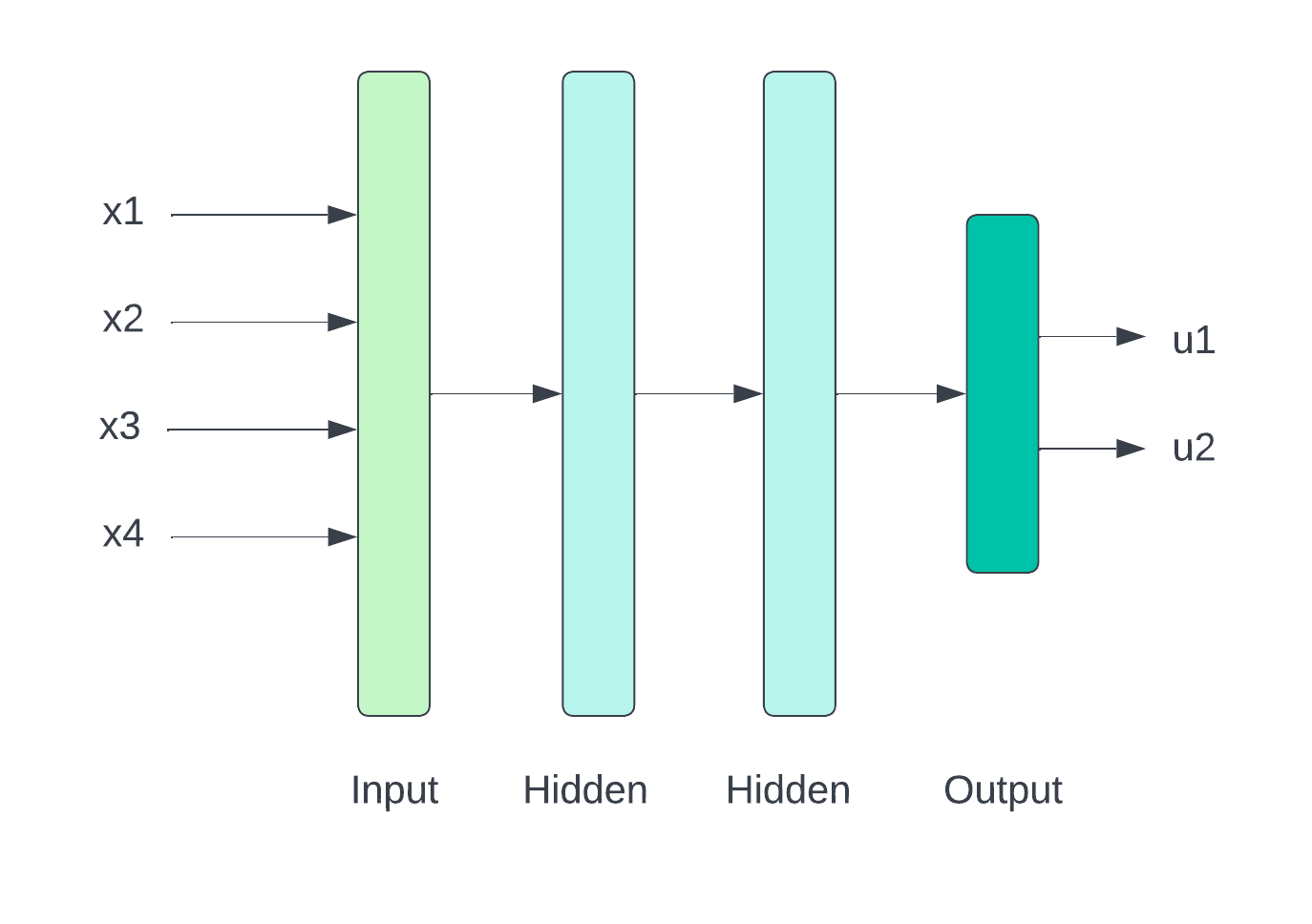}
    \centering
    \caption{Neural architecture}
    \label{fig:NNArchitecture}
\end{figure}

A simple feed-forward neural network with an input layer(4 states), two hidden layers(512 neurons each), and an output layer(2 controls) shown in Fig \ref{fig:NNArchitecture} is used for this study. 

\subsection{Behavioral Cloning}
Three different-sized data sets with 400k, 800k, and 1400k data points each were used for the experiments. The data points were produced by utilizing MPC simulations. As a result, 1600, 3200, and 5600 simulations total were produced for the three data sets, correspondingly. All experiments make use of the aforementioned neural architecture. The model is trained for 5 epochs. Experiments were also run by perturbing the states of the 800k and 1400k data sets using Gaussian noise.
% Plot for 400k data set
\begin{figure}[H]
    \begin{minipage}[c]{.48\linewidth}
        \centering
        \includegraphics[width=1.0\textwidth]{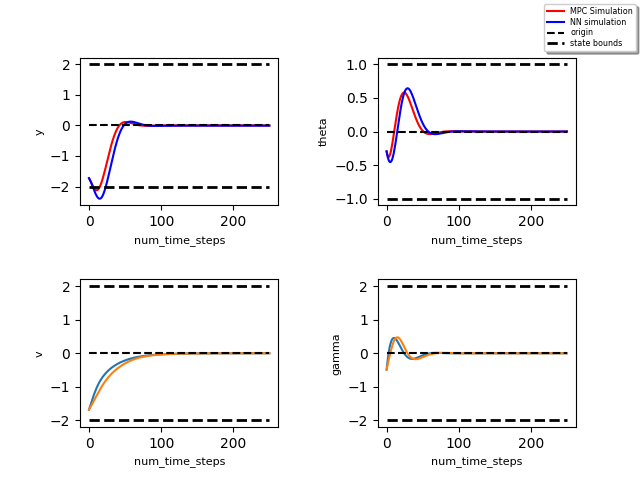}
         \subcaption{Inside Bound}
         \label{fig:bc4lib}
    \end{minipage}
    \hfill%
    \begin{minipage}[c]{.48\linewidth}
        \centering
        \includegraphics[width=1.0\textwidth]{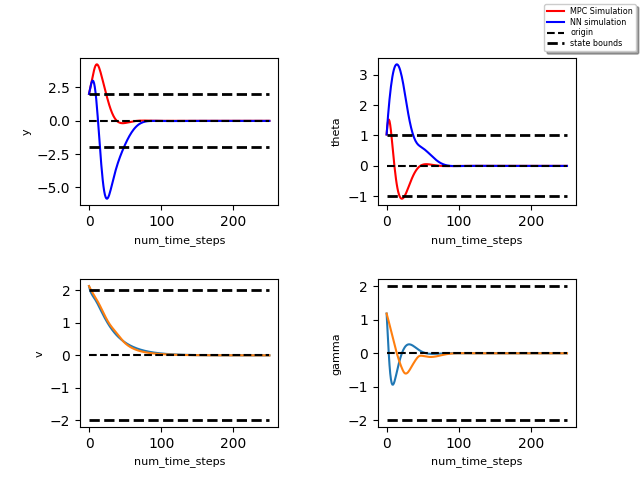}
        \subcaption{Outside Bound}
        \label{fig:bc4lob}
    \end{minipage}
\caption{MPC (vs) Neural state plot for 400K dataset}
\end{figure}

The above figures \ref{fig:bc4lib} and \ref{fig:bc4lob} show that the model trained on 400k data points shows a nearly perfect on the trajectories starting within the bounds but performs poorly on trajectories starting outside the bound. To make the model perform better for trajectories starting outside the bound we double the train data and generate predictions.

% Plot for 800k data set
\begin{figure}[ht]
    \begin{minipage}[c]{.48\linewidth}
        \centering
        \includegraphics[width=1.0\textwidth]{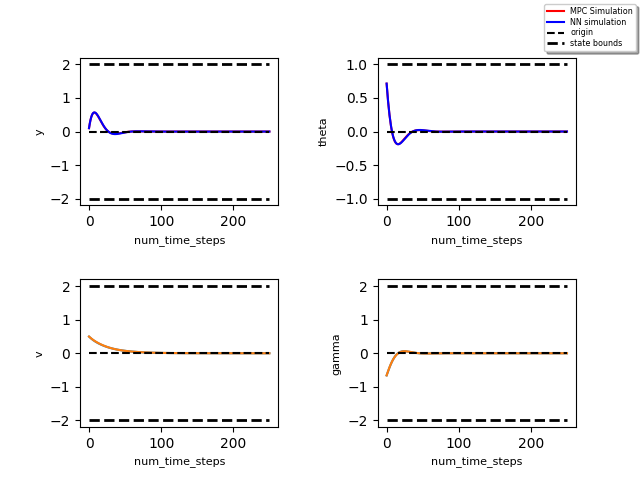}
         \subcaption{Inside Bound}
         \label{fig:bc4lib2}
    \end{minipage}
    \hfill%
    \begin{minipage}[c]{.48\linewidth}
        \centering
        \includegraphics[width=1.0\textwidth]{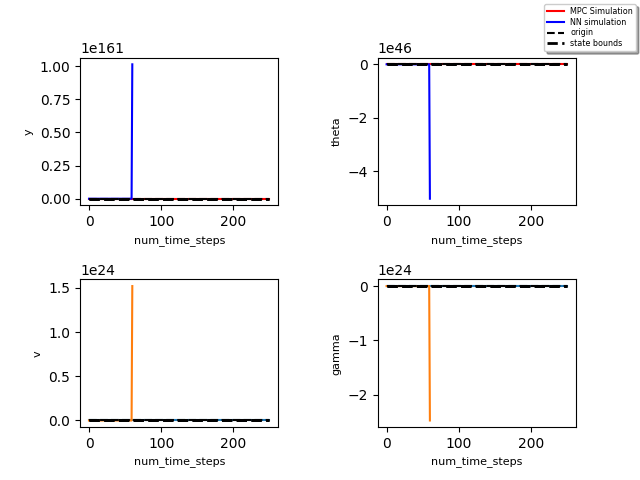}
        \subcaption{Outside Bound}
        \label{fig:bc4lob2}
    \end{minipage}
\caption{MPC (vs) Neural state plot for 800K dataset}
\end{figure}

Figures \ref{fig:bc4lib2} and \ref{fig:bc4lob2} show that the model works exactly as that of an MPC for trajectories starting inside the bound but explodes on the trajectories starting outside the bounds. This shows that the model is overfitting very much on the train data and is not able to generalize well on the region outside the given bounds. For making the model generalize well for regions outside the bounds we perturb the states of the dataset and generate predictions. 

% Plot for perturbed dataset
\begin{figure}[H]
    \begin{minipage}[c]{.48\linewidth}
        \centering
        \includegraphics[width=1.0\textwidth]{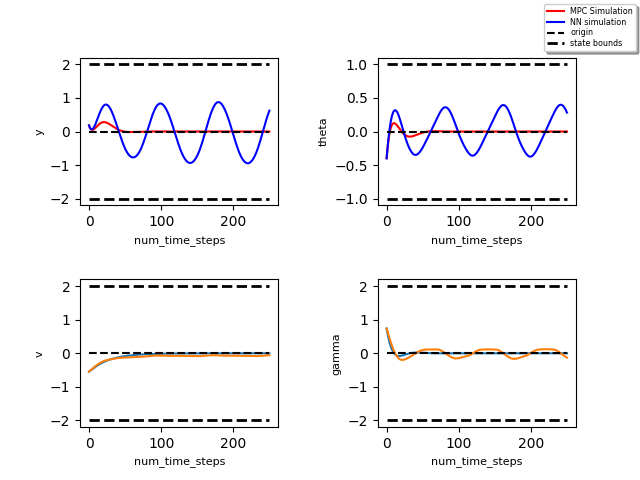}
         \subcaption{800K dataset}
         \label{fig:pert800}
    \end{minipage}
    \hfill%
    \begin{minipage}[c]{.48\linewidth}
        \centering
        \includegraphics[width=1.0\textwidth]{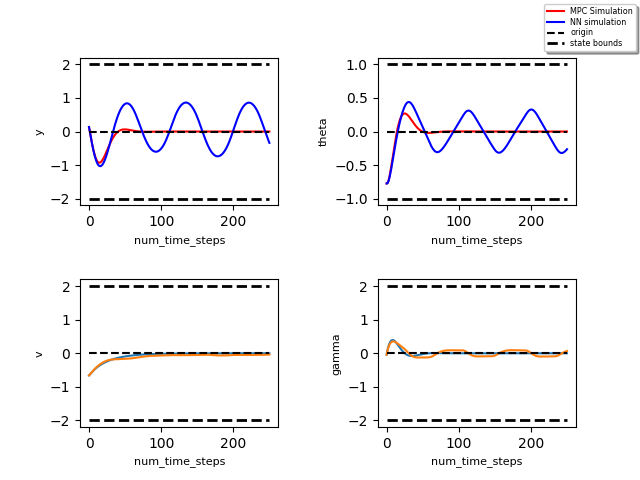}
        \subcaption{1400K dataset}
        \label{fig:pert1400}
    \end{minipage}
\caption{MPC (vs) Neural state plot for perturbed datasets}
\end{figure}

Figure \ref{fig:pert800} shows predictions generated on the model trained on the perturbed 800k dataset. We could see that the model fails to converge and keeps oscillating around stability. So, we increase the amount of perturbed data and train a model on the perturbed 1400k dataset. But predictions generated as in fig \ref{fig:pert1400} show that the model still fails to converge.  

\subsection{Dagger Approach}
The dagger approach starts with an initial dataset \(D_0\) having 40k data points in the training set and 10k data points in the validation set. A total of 10 iterations of the dagger are performed with each iteration adding 8k train and 2k validation samples. The neural network depicted by Fig \ref{fig:NNArchitecture} is used for all the iterations where the model is trained for 5 epochs each time. 
% 0th Iteration
\begin{figure}[H]
    \begin{minipage}[c]{.48\linewidth}
        \centering
        \includegraphics[width=1.0\textwidth]{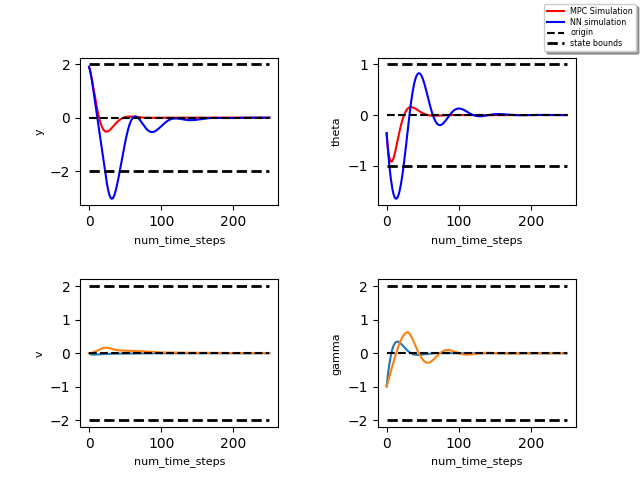}
         \subcaption{Inside Bound}
         \label{fig:iter0ib}
    \end{minipage}
    \hfill%
    \begin{minipage}[c]{.48\linewidth}
        \centering
        \includegraphics[width=1.0\textwidth]{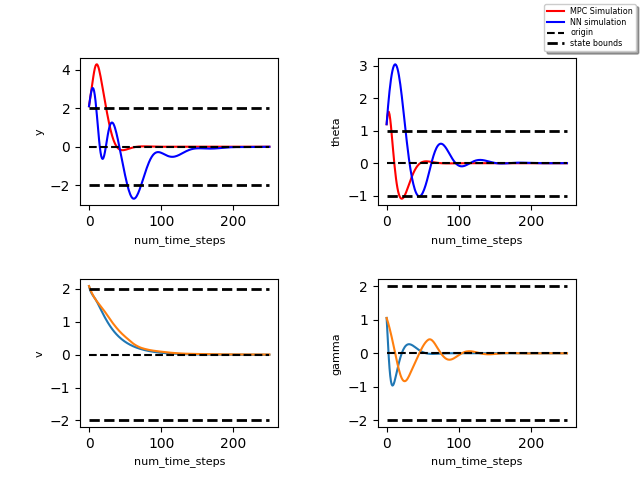}
        \subcaption{Outside Bound}
        \label{fig:iter0ob}
    \end{minipage}
    \caption{MPC (vs) Neural state plot for Iteration 0}
\end{figure}
% 5th Iteration
\begin{figure}[ht]
    \begin{minipage}[c]{.48\linewidth}
        \centering
        \includegraphics[width=1.0\textwidth]{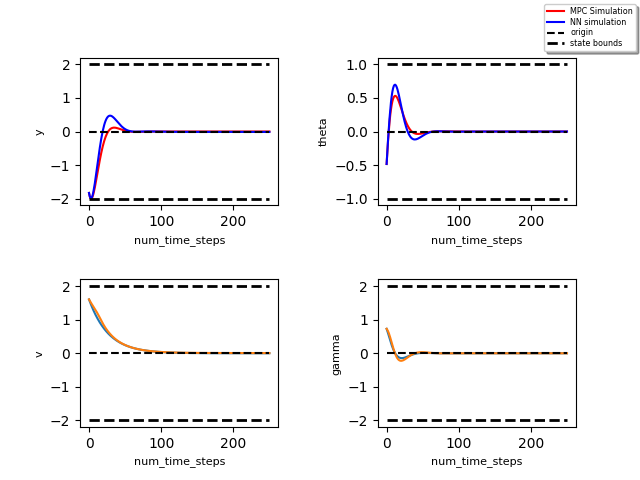}
         \subcaption{Inside Bound}
         \label{fig:iter5ib}
    \end{minipage}
    \hfill%
    \begin{minipage}[c]{.48\linewidth}
        \centering
        \includegraphics[width=1.0\textwidth]{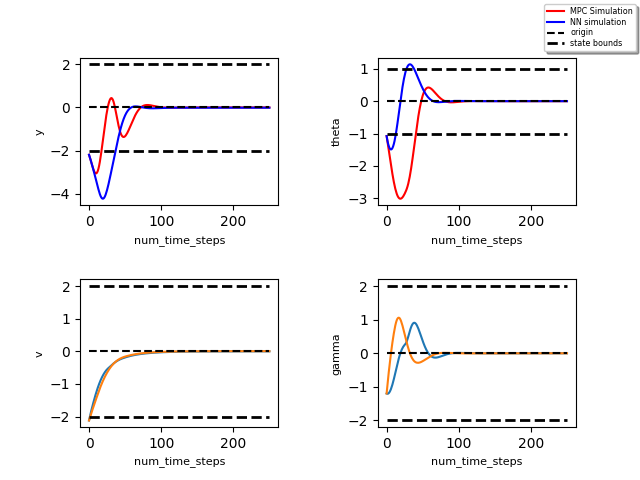}
        \subcaption{Outside Bound}
        \label{fig:iter5ob}
    \end{minipage}
    \caption{MPC (vs) Neural state plot for Iteration 5}
\end{figure}
% 10th iteration
\begin{figure}[ht]
    \begin{minipage}[c]{.48\linewidth}
        \centering
        \includegraphics[width=1.0\textwidth]{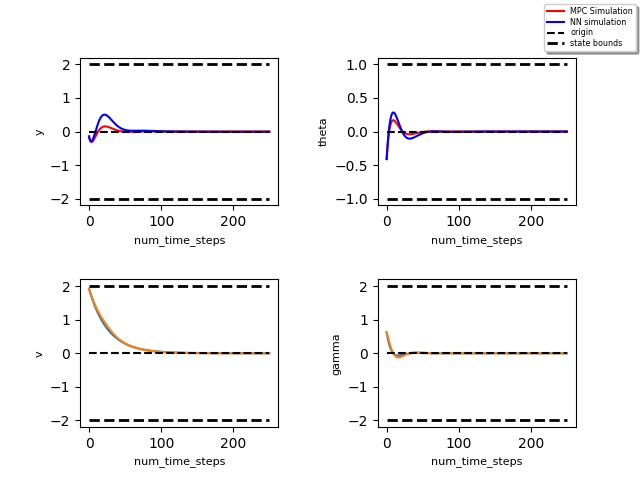}
         \subcaption{Inside Bound}
         \label{fig:iter10ib}
    \end{minipage}
    \hfill%
    \begin{minipage}[c]{.48\linewidth}
        \centering
        \includegraphics[width=1.0\textwidth]{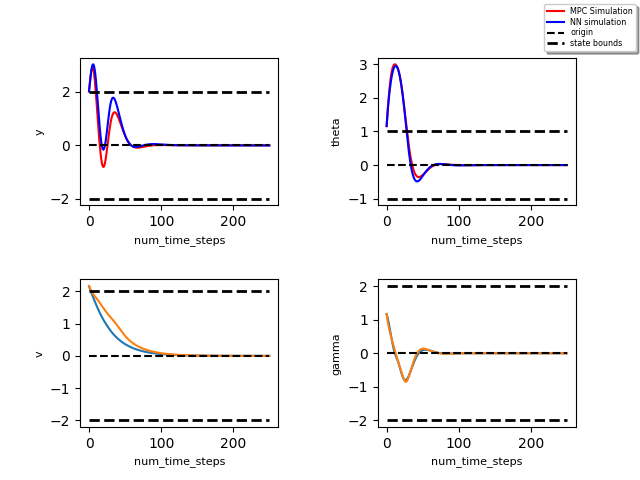}
        \subcaption{Outside Bound}
        \label{fig:iter10ob}
    \end{minipage}
    \caption{MPC (vs) Neural state plot for Iteration 10}
\end{figure}

From the above figures \ref{fig:iter0ib}, \ref{fig:iter0ob}, \ref{fig:iter5ib}, \ref{fig:iter5ob}, \ref{fig:iter10ib} and \ref{fig:iter10ob} we can see that as dagger iterations proceed, the model not only performs better on the trajectories starting inside the bounds but also generalizes very well on the trajectories starting outside the bounds.

\subsubsection{Loss Function}

The performance of the models is measured using the Root Mean Squared error (RMSE) between the actual trajectory computed using the MPC and the predicted trajectory using the neural network. For a bicycle robot since there are two predictions \( u_1 \) and \( u_2\) the RMSE is calculated as follows: 

\begin{equation}
    loss = \sqrt{ \frac{\sum_{i=1}^n \sum_{j=1}^2 (true\_u_{ij} - pred\_u_{ij})^2} {2n}  }
\end{equation}

Table \ref{table1} below shows the RMSE values for the different dagger iterations. 200 simulations were run in total and each simulation consists of 250 time steps. From the table, we can see that the error minimizes as the iterations proceed. The increase in error during iteration 2 is because the model explores previously unseen areas due to the compounding error effect.

\begin{table}[H]
\centering
\caption{RMSE values for different iterations}
\begin{tabular}{| l | l | l |}
\hline
\textbf{S.No} & \textbf{Dagger Iteration} & \textbf{RMSE} \\
\hline
1. & Iter 0 & 0.3208 \\
\hline
2. & Iter 2 & 0.4332 \\
\hline
3. & Iter 4 & 0.2552 \\
\hline
4. & Iter 6 & 0.2539 \\
\hline
5. & Iter 8 & 0.2441 \\
\hline
6. & Iter 10 & 0.2324 \\
\hline
\end{tabular}
\label{table1}
\end{table}

% \newpage
\bibliographystyle{unsrtnat}
\bibliography{references} 





% \end{thebibliography}

\end{document}